%% file: main.tex
\documentclass[letterpaper, amsmath, amssymb, 12pt]{revtex4-2}  

\usepackage{setspace}
\usepackage[a4paper, total={6in, 9in}]{geometry}
\usepackage[T1]{fontenc}
\usepackage{titlesec}
\usepackage{array}
\usepackage{siunitx}
\usepackage[normalem]{ulem}
\usepackage{xcolor}
\usepackage{booktabs}
\usepackage{multirow}
\usepackage{comment}

\usepackage{graphicx}
\usepackage{subcaption}
\usepackage[figurename=Figure, labelsep=period, font=scriptsize, labelfont=bf, format=plain]{caption}
\usepackage{placeins}
\usepackage{float}

\usepackage{bm}
\usepackage{mathrsfs}
\usepackage{exscale}
\usepackage{easybmat}
\usepackage{mathtools}

\usepackage{hyperref}
\usepackage{cleveref}
\hypersetup{
    colorlinks=true,
    linkcolor=blue,
    citecolor=blue,
    urlcolor=blue
}

\renewcommand\thesection{\arabic{section}}
\titleformat{\section}
  {\normalfont\bfseries\raggedright}{\thesection.}{1em}{}
\renewcommand\thesubsection{\thesection.\arabic{subsection}}
\titleformat{\subsection}
  {\normalfont\bfseries\raggedright}{\thesubsection}{1em}{}
\renewcommand\thesubsubsection{\thesubsection.\arabic{subsubsection}}
\titleformat{\subsubsection}
  {\normalfont\bfseries\raggedright}{\thesubsubsection}{1em}{}

\begin{document}

\title{Optimizing Flamelet Generated Manifold Models: A Machine Learning Performance Study}

\author{Reza Lotfi Navaei$^{1}$}
\author{Mohammad Safarzadeh$^{1}$}
 \author{Seyed Mohammad Jafar Sobhani$^{1,}$}\altaffiliation[Corresponding author ]{(smj.sobhani@modares.ac.ir)}

\affiliation{\vspace{2ex} \makebox[\linewidth][c]{{\normalfont $^1$}\footnotesize{Faculty of Mechanical Engineering, Tarbiat Modares University, P.O. Box 14115-111, Tehran, Iran}}\\
}

\begin{abstract}
In chemistry tabulations and Flamelet combustion models, the Flamelet Generated Manifold (FGM) is recognized for its precision and physical representation. The practical implementation of FGM requires a significant allocation of memory resources. FGM libraries are developed specifically for a specific fuel and subsequently utilized for all numerical problems using machine learning techniques. This research aims to develop libraries of Laminar FGM utilizing machine learning algorithms for application in combustion simulations of methane fuel. This study employs four Machine Learning algorithms to regenerate Flamelet libraries, based on an understanding of data sources, techniques, and data-driven concepts. 1. Multi-Layer Perceptron; 2. Random Forest; 3. Linear Regression; 4. Support Vector Machine. Seven libraries were identified as appropriate for constructing a database for training machine learning models, giving an error rate of 2.30\%. The default architectures of each method were evaluated to determine the optimal approach, leading to the selection of the MLP method as the primary choice. The method was enhanced through hyperparameter tuning to improve accuracy. The quantity of hidden layers and neurons significantly influences method performance. The optimal model, comprising four hidden layers with 10, 15, 20, and 25 neurons respectively, achieved an accuracy of 99.81\%.


\end{abstract}

\maketitle

\begin{table}[h!]
\centering
\begin{minipage}[t]{0.45\textwidth}
\section*{Nomenclature}
\renewcommand{\arraystretch}{1.2}
\begin{tabular}{ll}
\hline
$\rho$ & Density (kg\,m$^{-3}$) \\
$\tau$ & Time \\
$h$ & Enthalpy \\
$\dot{\omega}_F$ & Reaction rate (kg\,m$^{-3}$\,s$^{-1}$) \\
$\chi$ & Scalar dissipation rate (s$^{-1}$) \\
$T$ & Temperature \\
$Q_R$ & Radiation heat loss \\
$c_p$ & Specific heat capacity (J\,kg$^{-1}$\,K$^{-1}$) \\
$Da$ & Damköhler number \\
$D_z$ & Diffusion coefficient \\
$Le$ & Lewis number \\
$Z$ & Mixture fraction \\
\hline
\end{tabular}
\end{minipage}
\hfill
\begin{minipage}[t]{0.45\textwidth}
\section*{Acronyms}
\renewcommand{\arraystretch}{1.2}
\begin{tabular}{ll}
\hline
ANN & Artificial Neural Network \\
DNS & Direct Numerical Simulation \\
DT & Decision Tree \\
LR & Linear Regression \\
ML & Machine Learning \\
MLP & Multilayer Perceptron \\
PCA & Principal Component Analysis \\
RF & Random Forest \\
SGD & Stochastic Gradient Descent \\
SVM & Support Vector Machine \\
\hline
\end{tabular}
\end{minipage}
\end{table}

\section{Introduction}

Using machine learning (ML), many combustion challenges can be addressed, including reduced-order modeling, data processing, optimization, and control. Due to the increasing prevalence of data-driven methods, fluid mechanics can benefit from learning algorithms and present challenges that may encourage these algorithms to complement human understanding and engineering intuition \cite{compais_detection_2023, amiri_shear-thinning_2021}. 

Numerical modeling of combustion systems that incorporate intricate kinetic mechanisms presents a significant level of complexity. The quantity of species pertaining to light hydrocarbon fuels can extend to several hundred, while the chemical kinetic mechanism associated with certain fuels such as kerosene encompasses a few hundred to a few thousand reactions \cite{el_helou_comparison_2023}. The evolution of various species is represented by partial differential equations, which are coupled together. Additionally, the presence of turbulent flames introduces a diverse set of temporal and spatial dimensions, thereby adding complexity to the computational simulations. Additionally, the presence of turbulent flames introduces a diverse set of temporal and spatial dimensions, adding complexity to computational simulations \cite{mehrdad_bathaei_combustion_2023}. The temporal times of chemical processes involving intermediate species, specifically $H$, $CH_3$, and $C_2H_2$, are comparatively briefer in comparison to the time scales associated with $CO_2$ and $H_2O$ \cite{pasik_monoterpene_2025}. In spite of significant advances in supercomputers, simplification of the combustion process is still not feasible. Therefore, it is of great interest to develop more advanced models \cite{jafari_conceptual_2022}.

Recently, Brunton et al. \cite{brunton_machine_2020} reviewed the history, current developments, and emerging opportunities of machine learning for fluid mechanics, and Duraisamy et al. \cite{duraisamy_turbulence_2019} reviewed data-driven models constructed using machine learning to model turbulence. Numerous noteworthy studies have focused on the intersection of combustion and machine learning, offering a fresh outlook on the field of combustion research. In a previous study, Kalogirou et al. \cite{kalogirou_artificial_2003} presented an in-depth review of artificial intelligence techniques employed in the control and modeling of combustion processes. The study primarily emphasized the utilization of traditional methodologies, including expert systems, genetic algorithms, and neural networks \cite{amiri_deep-learning-enhanced_2025,afsharfard_modifying_2023}.

The recent advancements in ML have led to the emergence of various innovative and resilient data-driven models \cite{khaniki2025vision}. Consequently, it is imperative to conduct a comprehensive examination of the most recent combustion research, considering the utilization of state-of-the-art machine learning models. Ding et al. \cite{ding_machine_2021} used machine learning to identify turbulent combustion thermochemistry in equilibrated low swirl burners using methane, and Seltz et al. \cite{seltz_direct_2019} defined a new modeling framework to resolve the transport equation. To identify the critical role of mixture fraction in the process and to identify essential radicals involved in branching reactions and soot formation, Zdybal et al. \cite{zdybal_local_2023} analyzed data obtained in a 3D temporally evolving DNS of an n-heptane turbulent jet in the air. To predict soot production, Kessler et al. \cite{kessler_comparison_2021} compared neural networks, graph networks, and multivariate equations. Using ANN, Nguyen et al. \cite{nguyen_machine_2021} calculated combustion chemistry, temperature, radicals, intermediate species, and all stages of evolution (mixing, combustion, equilibrium), including radicals, $OH$, $O$, and $H_2O_2$.

A flamelet-based method, flamelet generated manifold (FGM), describes turbulent flames as ensembles of representative one-dimensional flamelets. In the FGM model, the flamelets are pre-computed and subsequently stored. A low-dimensional manifold is used to calculate the diffusion and convection effects of 1-D laminar flames \cite{joe_combustion_2023}. Control variables (CVs) space is created by transforming from the space-time domain ($x-t$) to mixture fraction and progress variable \cite{bao_large-eddy_2023}. The connection between the tabulated data and the flow field is established by incorporating either all the control variables or a selected subset of them. The implementation of this preprocessing technique leads to a reduction in computational expenses without compromising the accuracy of the results. Large mechanisms can be adopted by employing rigid formulations that are not reliant on fluid dynamic calculations.

Flamelet-based models use ML techniques to model complex non-linear data. ML methods require less memory compared to the FGM model because they only need to store the architecture characteristics and parameters \cite{ihme_combustion_2022}. Zhang et al. \cite{j_zhang_artificial_2020} effectively utilized ANN to model the flame of a methane-air mixture burner using FGM model. The efficacy of this methodology was evaluated by conducting a comparative analysis between the outcomes of numerical simulations and empirical observations. The utilization of ANNs in conjunction with flamelet-based models has been found to yield notable reductions in memory usage. Nonetheless, it is important to note that these models possess limited representation capabilities and necessitate extensive training durations, as mentioned in the study of different studies \cite{hansinger_deep_2022, hasani2025design} . In research conducted by Ihme et al. \cite{ihme_optimal_2009} a computational fluid dynamics (CFD) investigation was carried out on burner fueled by combination of methane and hydrogen. Instead of utilizing flamelet-progress variable (FPV) look-up tables, the researchers opted for the implementation of classical ANN. To enhance the efficiency of the neural network architecture, the researchers proposed the utilization of an optimization technique to determine the optimal number of layers and neurons.

Several optimization methods were used to optimize engine control parameters, including least squares support vector machines (LS-SVM), particle swarm optimization (PSO), and genetic algorithms (GA). A comparison of the GA-SVR model's performance with standard SVR models, random forest models, and back propagation neural network models was performed by Guo et al. \cite{guo_method_2022} Based on the results, it can be concluded that GA-SVR is more accurate and generalizable.

Advantageous in terms of memory performance compared to the complete replacement of look-up tables. However, the integration of ML models with look-up tables can pose challenges in both numerical implementation and practical usage. Bissantz et al. \cite{bissantz_application_2023} employed a dense neural network (DNN) to substitute table-based outcomes and a sparse principal component analysis (SPCA) to ascertain control variables (CVs) in their computational fluid dynamics (CFD) simulation of flame quenching in premixed methane-air systems. The FGM-DNN model incorporated a progress variable and temperature as control variables (CVs). Although the runtime and prediction accuracy exhibit comparable performance, a significant reduction in memory usage of up to 98\% can be achieved. The performance of the model is significantly impacted by both computer hardware and DNN architecture. In their study, Mousemi et al. \cite{mousemi_application_2023} employed a decision tree (DT) and an ANN model in conjunction with a tabulated chemistry approach to replicate premixed methane-air flames in a Bladed Ring Clearance burner. The utilization of DT and ANN models resulted in a reduction of memory requirements by 40\% and 92\% correspondingly. In addition, the decision tree model exhibited superior computational efficiency. However, it should be noted that DT models are susceptible to the issues of overfitting and entrapment.

RF and DT models are ML techniques that have the potential to reduce training time and reveal intricate relationships among parameters in combustion systems \cite{li_combining_2023}. Based on prior research, it has been observed that DT exhibit superior performance compared to RF owing to their greater flexibility in parameter adjustment \cite{saravanan_thermal_2023}. Recent research has also examined their application in combustion modeling. With RF and ANN, Ren et al. \cite{ren_lower-dimensional_2020} measured flame structure properties in which the RF model outperformed ANN in three DNS cases with different turbulence intensities. Natural gas engine output parameters were forecast using RF and ANN models by Liu et al. \cite{liu_steady-state_2022}. Both methods perform comparably, but ANN models can be challenging to tune because of their complex parameters. In their first attempt to predict non-linearity between combustion parameters, Yao et al. \cite{yao_gradient_2022} employed ensemble models to investigate the relationship between input variables, such as species concentrations, pressure, and temperature, and output variables, namely species compositions at subsequent times. The study revealed that DT exhibited superior performance compared RF in predicting species compositions and temperatures, on par with ANN.

According to research on machine learning in fluid mechanics and combustion, much of the research has focused on using data extracted from classical and traditional methods. ML models each have advantages and disadvantages. ANN models, for example, have longer training and more parameters to tune \cite{kartal_prediction_2022}. Overfitting and local optimum can happen with DTs \cite{owoyele_application_2020}. Models such as RF and LR exhibit a relatively lower level of complexity during the training process, albeit at the cost of reduced accuracy and increased memory consumption \cite{yang_application_2022}. In this study, four different ML methods are compared in FGM simulations. The optimum model is tuned using different methods to obtain the minimum reliable accuracy for the present study. Optimal parameters are determined by comparing the models.

\section{Governing Equations}

Combustion is a mass and energy conversion process in which chemical energy is converted into thermal energy. Fuel combines with atmospheric oxygen to generate carbon dioxide and water vapor, which have a lower enthalpy of production. Combustion is the primary energy source for transportation, heating, and electricity generation. The main goals of combustion research are to reduce emissions and increase the efficiency of systems since our fossil fuel resources are limited. To simulate the temperature and combustion species, it is necessary to solve the combustion equations. In this section, the definitions of thermodynamic variables and then the governing equations are presented. The governing equations of the Flamelet model and machine learning are also given briefly.

Non-premixed combustion occurs when the fuel and the oxidizer are injected separately into the combustor and experience simultaneous mixing and burning. The Flamelet methodology utilized in non-premixed combustion is founded upon the concept of characterizing the turbulent flame as an assemblage of laminar flame constituents that are integrated within a turbulent flow and engage in mutual interactions. The flame's local structure at every point along the flame front is expected to exhibit similarities to a laminar Flamelet, with the influence of turbulence limited to the evolution of the front.

The prevailing consensus is that the Flamelet concept holds true within the high Damköhler number (Eq.\ref{eq:1}), where the chemical time scale $\tau_c$ is compared to the flow time scale $\tau_t$:

\begin{equation}\label{eq:1}
    Da = \frac{\tau_c}{\tau_t}
\end{equation}

In contrast to the turbulent time scale, a high Damköhler number ($Da$) denotes rapid chemical reactions. In this situation, the chemical reactions exhibit a rapid response to alterations in flow, thereby rendering unsteady effects negligible. Another requirement is that the thickness of the flame should be sufficiently small compared to the turbulent length scales in order to prevent vortices from disturbing the internal structure of the flame. Despite the rapid nature of the chemistry involved, specifically the presence of a thin reaction layer, the diffusive flame thickness is comparatively larger and can potentially be influenced by the flow conditions.

One possible approach involves performing a conversion from the physical spatial-temporal domain to a new domain where the mixture fraction is introduced as an additional independent variable. Consequently, the newly established coordinate system is closely associated with an iso-surface of the mixture fraction, specifically the stoichiometric mixture fraction denoted as \( Z_{st} \). The coordinates \( Z_2 \) and \( Z_3 \) are situated within this iso-surface. The application of a Crocco-type coordinate transformation is carried out, wherein the transformation rules are methodically employed to the governing equations pertaining to species and energy. At this juncture, it is postulated that the flame front exhibits a unidimensional behavior. The terms associated with flame front gradient, specifically \( Z_2 \) and \( Z_3 \) gradients, are disregarded relative to the normal flame gradients. The flamelet equations can be expressed as follows (Eqs.(\ref{eq:2}) and (\ref{eq:3})), by considering \( Le_i = 1 \) and negligible Soret effect.

\begin{equation}\label{eq:2}
\rho \frac{\partial Y_i}{\partial \tau} = 
\frac{\rho \chi}{2 Le_i} \frac{\partial^2 Y_i}{\partial Z^2}
+ \frac{1}{4} \left( \frac{1}{Le_i} - 1 \right)
\left[
\frac{\partial \rho \chi}{\partial Z}
+ \rho \chi \frac{c_p}{\lambda} \frac{\partial}{\partial Z} \left( \frac{\lambda}{c_p} \right)
\right]
\frac{\partial Y_i}{\partial Z}
+ \dot{\omega_i}
\end{equation}

\begin{equation}\label{eq:3}
\rho \frac{\partial T}{\partial \tau} = 
\frac{\rho \chi}{2} \frac{\partial^2 T}{\partial Z^2}
+ \frac{\rho \chi}{2 c_p} \frac{\partial c_p}{\partial Z} \frac{\partial T}{\partial Z}
+ \frac{\rho \chi}{2 c_p} \sum_{i=1}^{N} \left( \frac{c_{p i}}{Le_i} \frac{\partial Y_i}{\partial Z} \right) \frac{\partial T}{\partial Z}
- \frac{1}{c_p} \sum_{i=1}^{N} h_i \dot{\omega_i}
+ \frac{Q_R}{c_p}
\end{equation}

\noindent
Here, \( \tau \), \( T \), \( Q_R \), \( c_{p_i} \), \( h_i \), \( \dot{\omega}_i \), \( Y_i \), and \( Le_i = \frac{\lambda}{\rho D_{im} c_p} \), and \( D_{im} \) represent the time, temperature, radiative heat loss, $i$-th specific heat, enthalpy, net rate of production, mass fraction, Lewis number, and multicomponent ordinary diffusion coefficient, respectively. 

In these equations, \( \chi \) is the instantaneous scalar dissipation rate defined by:
\begin{equation}\label{eq:4}
\chi = 2 D_Z \left( \nabla Z \right)^2
\end{equation}

Eqs.~(\ref{eq:2}) and (\ref{eq:3}) are commonly simplified by assuming that certain terms can be neglected. Specifically, in Eq.~(\ref{eq:2}), the second term on the right-hand side is considered negligible. Similarly, in Eq.~(\ref{eq:3}), both the second and third terms on the right-hand side are assumed to have negligible impact. Given these simplifications, the equations can be expressed in the following manner:

\begin{equation}\label{eq:5}
\rho \frac{\partial Y_i}{\partial \tau} = 
\frac{\rho \chi}{2 Le_i} \frac{\partial^2 Y_i}{\partial Z^2}
+ \dot{\omega}_k
\end{equation}

\begin{equation}\label{eq:6}
\rho \frac{\partial T}{\partial \tau} = 
\frac{\rho \chi}{2} \frac{\partial^2 T}{\partial Z^2}
- \frac{1}{c_p} \sum_{i=1}^{N} h_i \dot{\omega}_i
+ \frac{Q_R}{c_p}
\end{equation}

For the energy Eq. (\ref{eq:6}), although the term that involves the $Z$-derivative of the heat capacity is sometimes retained.

By utilizing a chemical mechanism, establishing suitable boundary conditions, and incorporating scalar dissipation rate profiles \( \chi(Z) \), the flamelet equations (Eqs.~(\ref{eq:2}) and (\ref{eq:5}) for combustion, and Eqs.~(\ref{eq:3}) and (\ref{eq:6}) for species) can be numerically solved to derive equations for the determination of species mass fractions (Eq.~(\ref{eq:7})) and temperature distributions (Eq.~(\ref{eq:8})):

\begin{equation}\label{eq:7}
Y_i = Y_i(Z, \chi)
\end{equation}

\begin{equation}\label{eq:8}
T = T(Z, \chi)
\end{equation}

The fundamental principle of the flamelet approach is predicated upon the utilization of a variable transformation, denoted as \( (x, t) \rightarrow (Z, \chi) \). In the event that the dynamics and flame structure become decoupled, the comprehensive depiction of the temperature and species profiles' spatial and temporal evolutions can be achieved solely through the spatial and temporal evolution of \( Z \).

\section{Methods and Materials}

As machine learning (ML) methods mature and become more available, the best fit for specific problems can be assigned to supervised learning algorithms. Algorithms are chosen based on the size and nature of the data. Depending on the quantity of available training data, various principles of thumb for choosing proper ML methods have been presented in Table \ref{tab:ml_rules} The most common machine learning method in combustion is supervised learning. In supervised learning, machine learning models are inferred from labeled datasets that map inputs to outputs. Fitting thermodynamic response functions can be done using regression techniques on continuous outputs. Using supervised learning, model parameters are learned from data by minimizing a loss function (Eq. (\ref{eq:9})).

\input{Tables/table1}

\begin{equation}\label{eq:9}
\arg_{\theta \in \mathcal{P}} \max E\left(Y, f(X, \theta)\right)
\end{equation}

\noindent
Here, \( E \) denotes the error, \( X \) is the input, \( Y \) the output, and \( P \) represents all model parameters. The implicit dependence of the mapping function \( f \) on the model parameters \( \theta \) is investigated.

Model architecture, feature set, and optimization method for selecting model parameters differ between supervised learning algorithms. Throughout this article, four different methods are used to recreate flamelet libraries, which are briefly discussed below.

\subsection{Machine Learning methods applied}

\subsubsection{Linear Regression using Stochastic Gradient Descent}
\begin{figure}[h!]
\centering
\includegraphics[width=0.4\textwidth]{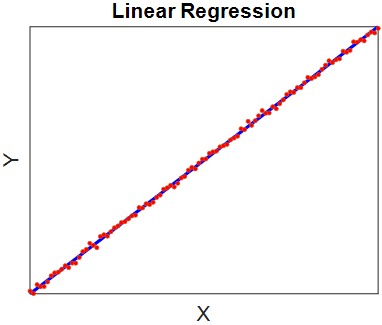} 
\caption{\footnotesize Linear Regression prediction}
\label{fig:lr_prediction}
\end{figure}

The stochastic gradient descent (SGD) method is a simple, yet efficient, approach to fitting linear classifiers and regressors with convex loss functions, such as support vector machines (linear) and logistic regressions. While linear regression (LR) (Figure \ref{fig:lr_prediction}) has existed in the machine learning community for a long time, it has recently gained significant attention in large-scale learning. SGD is not a machine learning model itself; rather, it is a method for training a model. It is merely a technique for optimization of linear models \cite{chander_hesitant_2023}. A linear regression learning routine is implemented, which supports a variety of loss functions and penalties for fitting linear regression equations. The model parameters \( \theta \) are determined by minimizing an error function (Eq. (\ref{eq:10})).

\begin{equation}\label{eq:10}
E = \frac{1}{n} \sum_{i=1}^{n} \left( Y_i - f(X_i, \theta) \right)^2
\end{equation}

\subsubsection{Decision Trees using Random Forest Regressor}
The decision tree is a model of possible outcomes that uses a flowchart-like structure to facilitate decision-making. Decision-tree algorithms are supervised learning algorithms. Categorical and continuous output variables can both be calculated using this method. Every decision tree has high variance, but when we combine them in parallel, the resultant variance is low because each decision tree is perfectly trained on the sample data. As a result, the output depends on more than one decision tree. For regression problems, the final output is the mean of all the outputs called Aggregation step. To form sample datasets for each model, Bootstrap is used to randomly sample rows and features from the dataset \cite{ma_developing_2023}.

\begin{figure}[h!]
\centering
\includegraphics[width=0.60\textwidth]{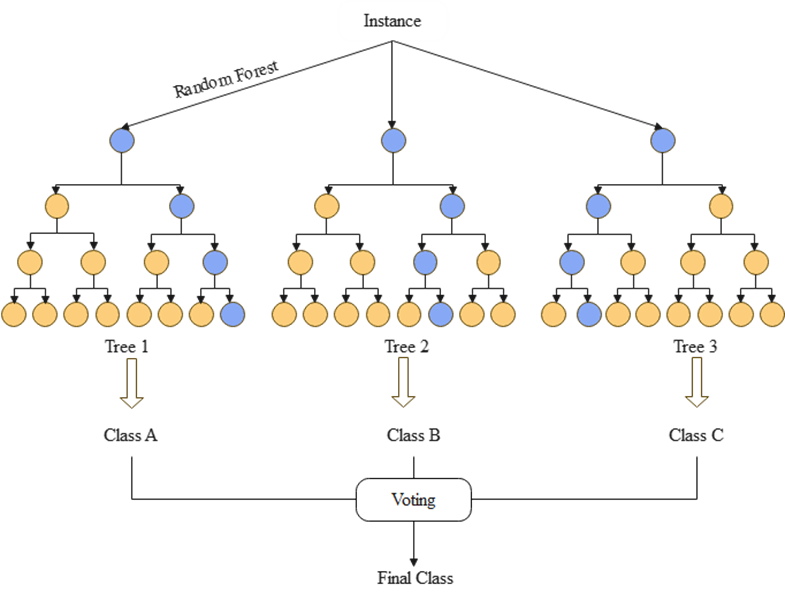} 
\caption{\footnotesize Schematic of a Random Forest algorithm with 3 decision trees}
\label{fig:random_forest}
\end{figure}

Using Bootstrap and Aggregation, also known as bagging, Random Forest is a method of performing regression and classification tasks. This involves combining multiple decision trees instead of relying solely on one decision tree to determine the final output(Figure \ref{fig:random_forest}). A Random Forest consists of multiple decision trees as its base learning model.

\subsubsection{Support Vector Machine}
\begin{figure}[h!]
\centering
\includegraphics[width=0.4\textwidth]{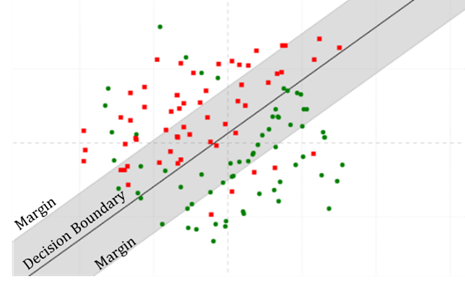} 
\caption{\footnotesize {SVM for a binary classification}}
\label{fig:svm_binary}
\end{figure}

The SVM is a deterministic algorithm that is used to create decision boundaries in datasets that are linearly separable. As depicted in Figure \ref{fig:svm_binary}, the decision boundaries are established by determining the utmost distance between points and the decision boundary, also known as support vector points. SVMs employ hyperplanes as decision boundaries for the purpose of data classification and regression \cite{manoharan_hybrid_2023}.

\begin{equation}\label{eq:11}
\sum_{i=1}^{M} (w_i X_i + b_i) = 0
\end{equation}

\noindent
Here, \( w \in \mathbb{R}^M \) represents the weights of the normal vector of the hyperplane, and \( b \) is the bias coefficient. The margins, which refer to the hyperplanes, can be mathematically represented as Eq. (\ref{eq:12}):

\begin{equation}\label{eq:12}
\sum_{i=1}^{M} (w_i X_i + b_i) = \pm 1
\end{equation}

\subsubsection{Neural Networks using Multi-Layer Perceptron (MLP)}
\begin{figure}[h!]
\centering
\begin{subfigure}{0.7\textwidth}
    \centering
    \includegraphics[width=\textwidth]{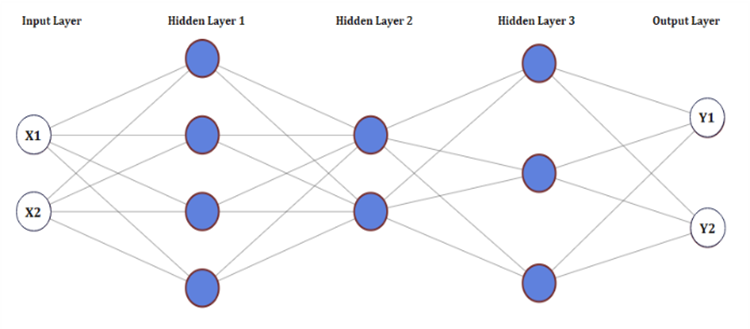} 
    \caption{}
\end{subfigure}

\vspace{1em}

\begin{subfigure}{0.4\textwidth}
    \centering
    \includegraphics[width=\textwidth]{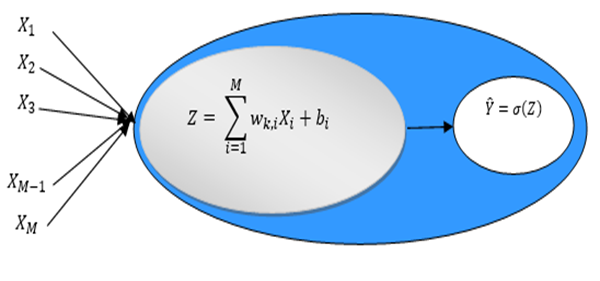} 
    \caption{}
\end{subfigure}

\caption{\footnotesize {Schematic of an MLP. (a) Network architecture, (b) Neuron and its operations.}}
\label{fig:mlp_schematic}
\end{figure}

Logistic regression cannot replicate non-linear decision boundaries, which limits its applicability to problems requiring complex physics-based classification. It is advisable to employ more dynamic algorithms for the applications. A Multi-Layer Perceptron (MLP), or fully connected feedforward neural network, is characterized by a configuration of neurons—also known as individual reasoning regression units—organized in a hierarchical arrangement of layers \cite{hemmat_prediction_2023} This structure is depicted in Figure \ref{fig:mlp_schematic}.

\begin{equation}\label{eq:13}
\hat{Y} = \sigma(Z) \quad \text{while} \quad Z = \sum_{i=1}^{M} w_{k,i} X_i + b_k
\end{equation}

\noindent
$\sigma$ is the neuron activation function. In a Multi-Layer Perceptron (MLP), the outputs of the previous layer are utilized as features for each neuron in the subsequent layer. MLP establishes a mapping from input to output. Researchers have developed specialized neural networks to address the needs of specific applications.

\subsection{Comparison of applied methods}

In this study, the aforementioned four methods have been employed for the purpose of predicting temperature and species mass fraction. Based on the investigations, each method has its own positive and negative points, which are mentioned in Table~\ref{tab:2}.

\input{Tables/table2}

In order to obtain high-accuracy data through the machine learning model for the present study, it is necessary to choose the most optimal method and then use different architectures to increase the model's accuracy so that the data satisfies the conditions of the problem well.

\subsection{Numerical Method}

In this research, the primary data are obtained by solving the Flamelet equations using two different approaches. The first approach is to solve the Flamelet equations with respect to the mixture fraction (\textit{Z}-flamelet) using FlameMaster, an open-source C++ program package for 0D combustion and 1D laminar flame calculations. The second approach involves solving 1D Flamelet equations using a CFD package. The data with higher accuracy, upon comparison, will be used for the subsequent processes \cite{smooke_comparison_1988}.
In both methods, a structured grid focused on the flame region is discretized using the central differencing method. The hardware used for this research includes a system equipped with an Intel Core i7-5820K 3.3~GHz CPU with 6 cores and 12 threads, 16~GB of DDR4 RAM, and 15~MB of cache.

\subsection{Algorithm configuration}
The output data format extracted is ``.fla'', which cannot be read directly in Python. For this reason, a code in C++ was developed to read this data and convert the file into a dataset in ``.csv'' format, organized with respect to the scalar dissipation rate. After transferring the data to Python, different machine learning models were evaluated, and an appropriate model was selected based on the type and range of the data.

\begin{figure}[h!]
\centering
\includegraphics[width=0.75\textwidth]{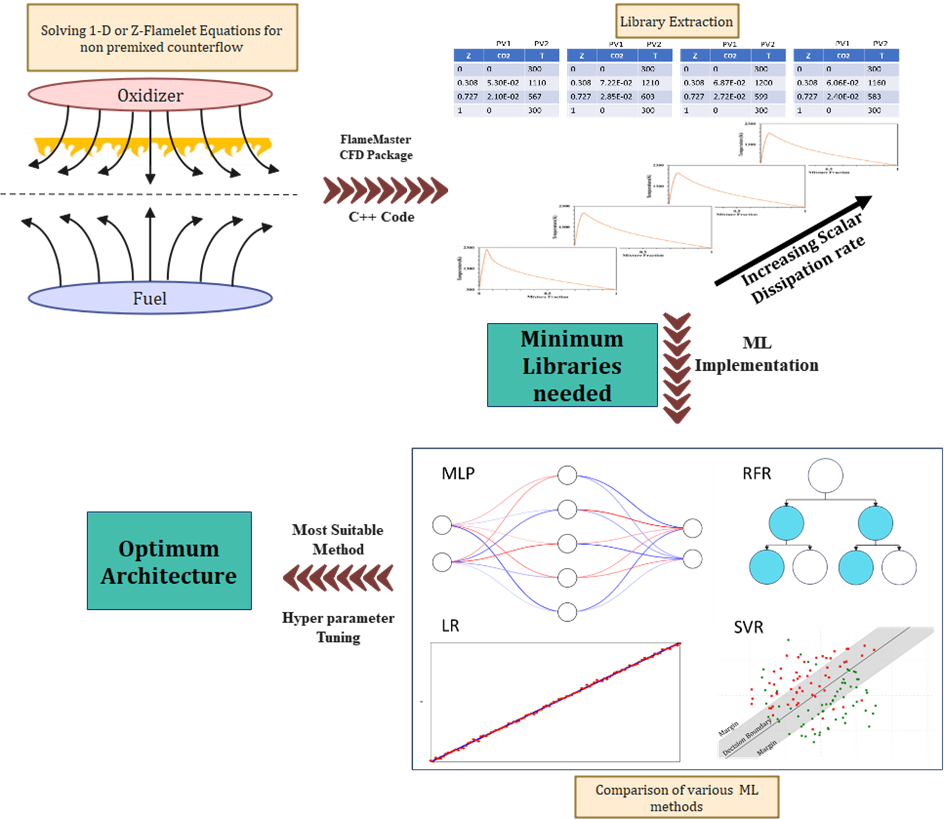} 
\caption{\footnotesize {Procedure flowchart of the present study.}}
\label{fig:flowchart_procedure}
\end{figure}

After choosing the optimal method, the output data from the method were compared with other data solved by Flamelet equations to validate the reproduced data, and maximum errors will be computed. In cases with low accuracy, the model's architecture is changed to satisfy the desired conditions for prediction. After satisfying all the conditions, the optimal model and its reports are extracted for further investigation (schematic shown in Figure \ref{fig:flowchart_procedure}).

\section{Results and Discussion}

To use machine learning in developing the flamelet combustion model library, first need to create a library using this method; so that the necessary data for inputting into the machine learning algorithm is provided. In creating the Flamelet model library, the conventional steady counterflow flame configuration is commonly used, and the library takes shape by changing the distances between fuel and oxidizer. Obviously, in the first step, the accuracy of the generated data needs to be verified. Therefore, in this section, the accuracy of the non-premixed flame is first validated, and then the library is formed by changing the scalar dissipation rates. Finally, these data are used and reproduced by various machine learning methods.

\subsection{Validation}

\subsubsection{Problem Definition}

\begin{figure}[h!]
\centering
\includegraphics[width=0.35\textwidth]{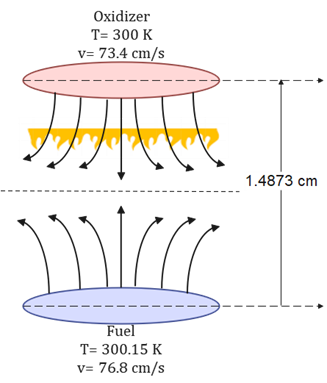} 
\caption{\footnotesize {Schematic of computational domain and boundary conditions}}
\label{fig:comp_domain}
\end{figure}

It is necessary to evaluate the accuracy of the modeling after examining the number of mesh grids required to construct a Flamelet library. Experimental and numerical results are presented for the cross-flow of methane fuel injected at a distance of 1.4876~cm with a velocity of 76.8~cm/s at a temperature of 300.15~K, and air injected with a velocity of 73.4~cm/s at a temperature of 300~K. Consequently, a one-dimensional cross-flow was modeled (Figure~\ref{fig:comp_domain}).

\subsubsection{Mesh Study}

To validate the model, the present simulation was compared with the experimental and numerical results of Smooke et al. \cite{smooke_comparison_1988}\ for counterflow methane combustion. This experiment reported the temperature and mass fractions of species such as CO$_2$ and OH at a pressure of 1~atm. First, the simulation results were validated by comparing them with those reported in Smooke et al.'s study \cite{smooke_comparison_1988}. 

Subsequently, the required library for the Flamelet method was prepared by solving the Flamelet equations for a non-premixed counterflow with various domain lengths. However, prior to forming the library, it was necessary to investigate the number of grid points required in one-dimensional space. Four grids containing 10, 30, 60, and 120 points were evaluated. The temperature and mass fractions of CO$_2$ and OH obtained from these four grids were compared.

\begin{figure}[h!]
\centering

\begin{subfigure}[b]{0.3\textwidth}
    \centering
    \includegraphics[width=\textwidth]{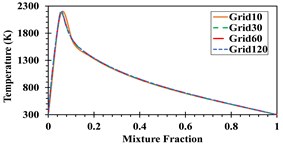} 
    \caption{}
\end{subfigure}
\hfill
\begin{subfigure}[b]{0.3\textwidth}
    \centering
    \includegraphics[width=\textwidth]{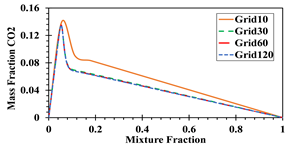} 
    \caption{}
\end{subfigure}
\hfill
\begin{subfigure}[b]{0.3\textwidth}
    \centering
    \includegraphics[width=\textwidth]{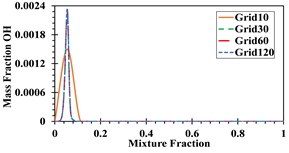} 
    \caption{}
\end{subfigure}

\caption{\footnotesize {Grid independency (a) Temperature (b) CO$_2$ (c) OH}}
\label{fig:grid_independency}
\end{figure}

As shown in Figure~\ref{fig:grid_independency}(a), the temperature profiles exhibit negligible differences across the four grids, indicating mesh independence. However, in the results for the mass fractions of CO$_2$ and OH shown in Figure~\ref{fig:grid_independency}(b) and \ref{fig:grid_independency}(c), the 30-, 60-, and 120-point grids yield consistent results, while the 10-point grid deviates significantly. Therefore, a 30-point grid was selected for use in the formation of the Flamelet library.

\subsubsection{Comparison of flamelet equation and experimental data}

\begin{figure}[h!]
\centering

\begin{subfigure}[b]{0.3\textwidth}
    \includegraphics[width=\textwidth]{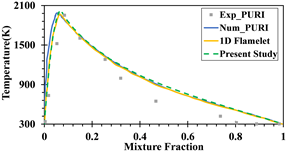}
    \caption{}
\end{subfigure}
\hfill
\begin{subfigure}[b]{0.3\textwidth}
    \includegraphics[width=\textwidth]{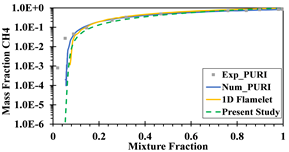}
    \caption{}
\end{subfigure}
\hfill
\begin{subfigure}[b]{0.3\textwidth}
    \includegraphics[width=\textwidth]{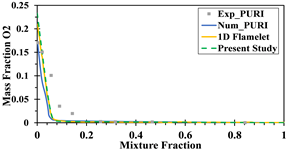}
    \caption{}
\end{subfigure}

\vspace{1em}

\begin{subfigure}[b]{0.3\textwidth}
    \includegraphics[width=\textwidth]{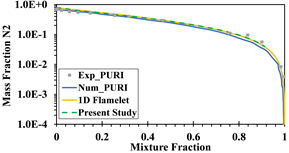}
    \caption{}
\end{subfigure}
\hfill
\begin{subfigure}[b]{0.3\textwidth}
    \includegraphics[width=\textwidth]{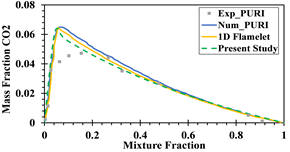}
    \caption{}
\end{subfigure}
\hfill
\begin{subfigure}[b]{0.3\textwidth}
    \includegraphics[width=\textwidth]{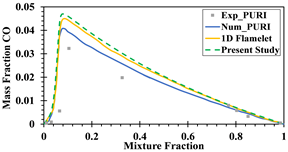}
    \caption{}
\end{subfigure}

\vspace{1em}

\begin{subfigure}{0.4\textwidth}
    \includegraphics[width=\textwidth]{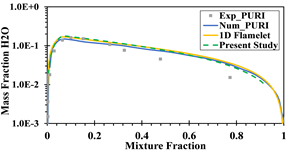}
    \caption{}
\end{subfigure}
\hfill
\begin{subfigure}{0.4\textwidth}
    \includegraphics[width=\textwidth]{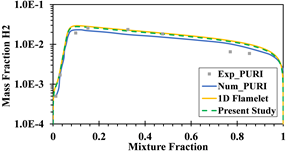}
    \caption{}
\end{subfigure}

\caption{\footnotesize {Comparison of numerical and experimental results of 1D Flamelet equation and this study. (a) Temperature (b) CH$_4$ (c) O$_2$ (d) N$_2$ (e) CO$_2$ (f) CO (g) H$_2$O (h) H$_2$.}}
\label{fig:comparison_3perrow}
\end{figure}

As shown in Figure~\ref{fig:comparison_3perrow}, the temperature results are consistent with the modeling results in \cite{smooke_comparison_1988}; however, they deviate slightly from the experimental observations. In the experimental results, the maximum temperature occurs at a fuel mass fraction of approximately 0.1, whereas in the modeling results, it is observed at a fuel mass fraction of about 0.07.

The results for species mass fractions show a more significant deviation from the experimental data compared to the temperature results, but they are more consistent with the simulation results (see Figure~\ref{fig:comparison_3perrow}). A slight difference is observed between the experimental data and the results of this study at a mixture fraction of 0.1. Regarding the species N$_2$, there is a strong agreement between the experimental and numerical results, with only minor differences observed.

Based on Figure~\ref{fig:comparison_3perrow}, it can be observed that the species mass fraction results for combustion products such as CO and CO$_2$ are predicted with less accuracy compared to the primary combustion species, namely methane and oxygen. Good agreement is also observed for H$_2$O and H$_2$ species with the numerical results. 
However, for the H$_2$O species, a deviation from the experimental data is noted for mixture fractions greater than 0.26. This deviation begins at a mixture fraction of 0.19 for 0.26 and eventually reaches 0.167 at a mixture fraction of 0.78. Similarly, a deviation occurs for the H$_2$ species at a mixture fraction of 0.77, where the value reaches 0.0066. 
Therefore, the present approach is capable of accurately predicting temperature and species mass fractions within the examined range.

\subsection{Minimum number of required libraries}

The second step investigated in this study is the independence of the machine learning model from the number of libraries used for training. To construct a dataset for training, various strategies and conditions were explored to determine the minimum required number of libraries that would reduce computational costs. For these models to be reliable, their accuracy with the least number of libraries must remain acceptable.

\begin{figure}[h!]
\centering

\begin{subfigure}[b]{0.3\textwidth}
    \centering
    \includegraphics[width=\textwidth]{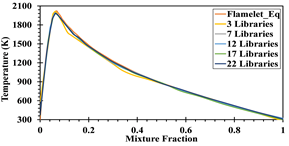} 
    \caption{}
\end{subfigure}
\hfill
\begin{subfigure}[b]{0.3\textwidth}
    \centering
    \includegraphics[width=\textwidth]{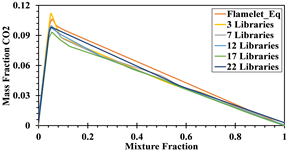} 
    \caption{}
\end{subfigure}
\hfill
\begin{subfigure}[b]{0.3\textwidth}
    \centering
    \includegraphics[width=\textwidth]{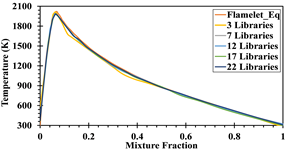} 
    \caption{}
\end{subfigure}

\caption{\footnotesize {Minimum library needed for forming dataset required for training machine learning models. (a) Temperature (b) CO$_2$ (c) Combined comparison}}
\label{fig:min_library_sidebyside}
\end{figure}

As shown in Figure~\ref{fig:min_library_sidebyside}, multiple models were developed using 3, 7, 12, 17, 22, and 27 libraries to predict temperature, CO$_2$, and OH species. Each library contained 250 data points. The results for temperature and the mass fractions of CO$_2$ and OH, corresponding to different numbers of libraries, are presented in Figure~9, where the scalar dissipation rate is 5~(1/s).

The findings indicate that all models produce an average error of less than 4\% for temperature and less than 15\% for CO$_2$, compared to the reference results extracted from the Flamelet equations, as summarized in Table~\ref{tab:3}. Moreover, increasing the number of libraries improves agreement with the Flamelet-based reference results. Notably, high prediction accuracy for temperature can be achieved with as few as three libraries. However, more libraries are needed for improved predictions of species mass fractions.

\input{Tables/table3}

\begin{figure}[h!]
\centering
\includegraphics[width=0.35\textwidth]{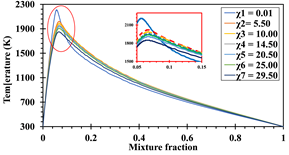} 
\caption{\footnotesize {Temperature difference for different scalar dissipation rates}}
\label{fig:temp_vs_scalar_dissipation}
\end{figure}

Considering the trade-off between error and computational cost, seven libraries were selected for further analysis. Accordingly, all methods were trained using seven Flamelet libraries with different scalar dissipation rates ($\chi$) ranging from 0.01 to 29.5~(1/s), as listed in Table~\ref{tab:4} and illustrated in Figure~\ref{fig:temp_vs_scalar_dissipation}.

\input{Tables/table4}

\subsection{Comparison of different ML algorithms}

Data from the Flamelet library were reproduced using various algorithms. Four different methods were employed, each representing a distinct modeling approach: a neural network via the Multi-Layer Perceptron (MLP) method, a linear model via Linear Regression (LR), a decision tree approach via Random Forest Regression (RFR), and a hyperplane-based model via Support Vector Regression (SVR). The predictive accuracy of these models varied significantly when applied to the initial dataset.

\begin{figure}[h!]
\centering

\begin{subfigure}[b]{0.45\textwidth}
    \centering
    \includegraphics[width=\textwidth]{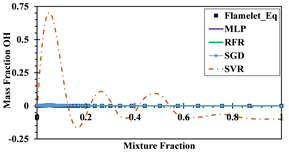} 
    \caption{}
\end{subfigure}
\hfill
\begin{subfigure}[b]{0.45\textwidth}
    \centering
    \includegraphics[width=\textwidth]{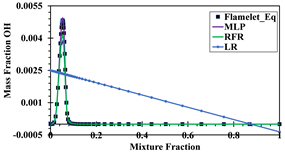} 
    \caption{}
\end{subfigure}

\caption{\footnotesize {Predicted data in different models (a) considering the SVR algorithm and (b) without considering the SVR algorithm}}
\label{fig:prediction_models}
\end{figure}

The SVR method exhibited high noise, large errors, and inconsistent predictions, particularly in reproducing the OH species mass fraction. As shown in Figure~\ref{fig:prediction_models}, SVR produced negative values in some cases, rendering its predictions unreliable. Consequently, the SVR method is excluded from further consideration in this study.

\begin{figure}[h!]
\centering

\begin{subfigure}[b]{0.3\textwidth}
    \includegraphics[width=\textwidth]{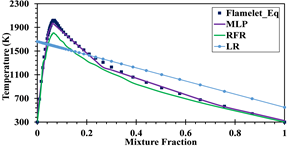}
    \caption{}
\end{subfigure}
\hfill
\begin{subfigure}[b]{0.3\textwidth}
    \includegraphics[width=\textwidth]{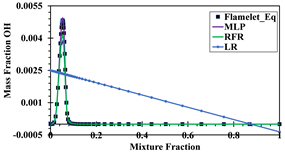}
    \caption{}
\end{subfigure}
\hfill
\begin{subfigure}[b]{0.3\textwidth}
    \includegraphics[width=\textwidth]{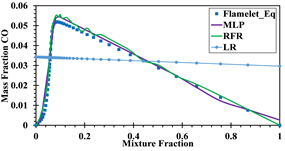}
    \caption{}
\end{subfigure}

\vspace{1em}

\begin{subfigure}[b]{0.3\textwidth}
    \includegraphics[width=\textwidth]{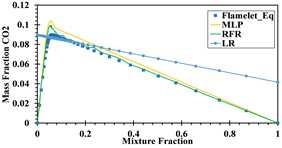}
    \caption{}
\end{subfigure}
\hfill
\begin{subfigure}[b]{0.3\textwidth}
    \includegraphics[width=\textwidth]{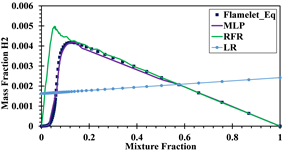}
    \caption{}
\end{subfigure}
\hfill
\begin{subfigure}[b]{0.3\textwidth}
    \includegraphics[width=\textwidth]{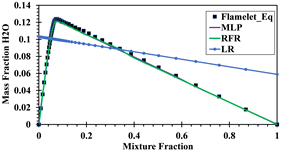}
    \caption{}
\end{subfigure}

\caption{\footnotesize {Comparison of the results of different ML methods (a) Temperature (b) Mass fraction of OH (c) Mass fraction of CO (d) Mass fraction of CO$_2$ (e) Mass fraction of H$_2$ (f) Mass fraction of H$_2$O.}}

\label{fig:ml_method_comparison}
\end{figure}

To evaluate the predictive performance for OH species, the MLP, RFR, and LR models were compared against the results from the Flamelet equations, as illustrated in Figure~\ref{fig:ml_method_comparison}. Among these, the MLP algorithm provided the most accurate and consistent predictions, both in value and trend. The prediction accuracies for OH mass fraction were 98.96\% for MLP, 68.89\% for RFR, and 14.44\% for LR. The performance of the LR model was especially poor, showing significant deviation and incorrect trend behavior.

The models also displayed varied performance when predicting other species. For CO mass fraction, the MLP, RFR, and LR models achieved accuracies of 87.71\%, 86.98\%, and 57.72\%, respectively. For CO$_2$, the models reached higher accuracies: 97.96\% for MLP, 91.51\% for RFR, and 80.90\% for LR. Despite acceptable accuracy in some cases, the LR model exhibited trend prediction failures and generated physically implausible outputs, thereby disqualifying it from further use.

In predicting the H$_2$O species, the MLP and RFR models showed strong agreement with the numerical results from the Flamelet equations, achieving prediction accuracies of 97.80\% and 98.78\%, respectively. In contrast, the LR model produced a significantly lower accuracy of 76.22\% and inconsistent behavior. The RFR model’s superior performance at low mixture fractions improved its average accuracy, but it still showed noise and inaccuracies in other regions.

The prediction of H$_2$ species mass fraction further highlighted the strengths of the MLP model. As shown in Figure~\ref{fig:ml_method_comparison}, the MLP model maintained superior trend fidelity and numerical accuracy, reaching a prediction accuracy of 93.35\%. The RFR model, with an average accuracy of 59.62\%, demonstrated considerable error, particularly for mixture fractions below 0.1 and above 0.1, where it exhibited noise. The LR model again performed the worst, with only 21.62\% accuracy and numerous erroneous predictions.

Based on these comprehensive evaluations, the MLP model was selected as the primary method for constructing the final predictive model using the Flamelet results. This approach was chosen for its high accuracy, reliable trend replication, and overall robustness, with the objective of maximizing model precision and minimizing prediction error.

\subsection{Different architectures of MLP}

Machine learning models provide different accuracies in predicting different phenomena, and this accuracy can vary depending on the nature of the data. In other words, one model may justify and predict a phenomenon in a way that another model cannot achieve the same level of accuracy. In addition to the overall difference in accuracy, the accuracy in each predicted parameter is also not the same. Suppose a model can achieve a general accuracy 
\[
\sqrt{\frac{\sum_{i=1}^{n} (\hat{Y}_i - Y_i)^2}{n}}
\]
of 96\%. The level of precision can be subject to fluctuations contingent upon the dataset's characteristics and the extent of numerical values involved.

Table~5 presents five distinct models utilizing varying architectures of the MLP algorithm, each exhibiting differing levels of accuracy. Each of the models mentioned above exhibits varying levels of accuracy when it comes to predicting temperature and species.
Table~\ref{tab:5} aims to determine the optimal level of prediction accuracy, thereby preventing researchers from further enhancing the precision of data for justification and reproducibility in subsequent studies.

\input{Tables/table5}

\begin{figure}[h!]
\centering

\begin{subfigure}[b]{0.3\textwidth}
    \includegraphics[width=\linewidth]{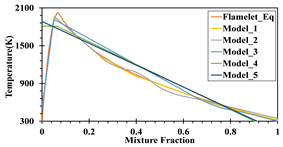}
    \caption{}
\end{subfigure}
\hfill
\begin{subfigure}[b]{0.3\textwidth}
    \includegraphics[width=\linewidth]{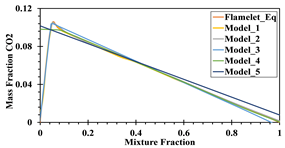}
    \caption{}
\end{subfigure}
\hfill
\begin{subfigure}[b]{0.3\textwidth}
    \includegraphics[width=\linewidth]{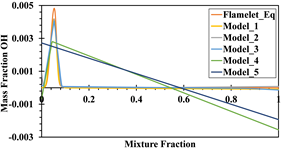}
    \caption{}
\end{subfigure}

\caption{\footnotesize Comparing the results of different models with different accuracies.}
\label{fig:models_3inrow}
\end{figure}

Figure~\ref{fig:models_3inrow} displays the output plots for the temperature, CO$_2$, and OH species models presented in Table~\ref{tab:5}. The figure indicates that Models~4 and~5 have deviated from the expected plot shape. The models exhibit a decreasing trend in the mixture fraction that is less than 0.1, despite the expectation for an increase in the parameters. Models~4 and~5 reproduced unacceptable results in forecasting OH mass fraction.

Upon analysis of the graphical representations of the five models mentioned in Figure~\ref{fig:models_3inrow}, it can be inferred that Table~\ref{tab:6} is valuable in comprehending the presented data. Table~6 was compiled by examining the temperature and CO$_2$ species in nine different mass fractions.

\input{Tables/table6}

\subsubsection{Uniform structure models}

After analyzing the specified methods in previous sections, it was determined that the MLP method exhibited better prediction accuracy and data generation capabilities, thereby establishing it as the optimal model for data generation. This model's notable characteristic is its capacity to make highly accurate predictions for nonlinear phenomena, which may be enhanced through architectural modifications. Changing the architecture of neural networks mostly refers to changes in the hidden layers and neurons. 

In order to select the optimal model with reliable performance, modifications were made to its architecture by increasing the number of hidden layers from 1 to 5 and varying the number of neurons (5, 10, 15, 20, 25). All presented models demonstrate uniform architectures, ensuring the number of neurons remains consistent across all hidden layers; since non-uniform architectures are more complex, they are more difficult to design and fine-tune.

The models presented in Table~\ref{tab:7} can be evaluated using accuracy and error classification, as demonstrated in Table~8. The MLP5-5 model is deemed inadequate for user modeling. The model's accuracy is below 50\%, and its error exceeds 0.01, making it unreliable for utilization in modeling. Conversely, models exhibiting an accuracy exceeding 91\% may be given precedence.

\input{Tables/table7}

The accuracies mentioned in Table~\ref{tab:7} are classified into five different groups according to their accuracy and their mean squared error (MSE). Moreover, in the case that the MSE error is below 0.005, the model may be deemed the optimal model. Only the MLP5-25, MLP5-15, and MLP4-20 models meet both criteria. The dissimilarity between the models mentioned above lies in their respective durations for training and prediction time. Thus, the MLP5-15 model is a viable option for presenting the findings. Additionally, Table~\ref{tab:8} compares the accuracy of other models used in this study with the MLP models.

\input{Tables/table8}

\subsection{Hyperparameter tuning}

As demonstrated in the preceding section, the inclusion of additional hidden layers and neurons has significantly improved the accuracy of the prediction results. Nevertheless, it is imperative to acknowledge that these parameters are not the only indicators of effectiveness, thus necessitating the investigation and assessment of additional parameters for MLP. This section examines hyperparameters, including the number of hidden layers and neurons, which significantly impact the accuracy of machine learning models in predicting data. By considering all effective factors in MLP algorithm accuracy, we can choose an optimal model based on hidden layers, neurons, activation function, solver, alpha coefficient, learning rate, and tolerance.

\begin{itemize}
    \item \textbf{Number of hidden layers:} In the initial form of MLP, a single hidden layer is considered. However, this study used this number from 1 to 5 to find the optimal hyperparameter.
    
    \item \textbf{Number of neurons in each layer:} This number is set to 100 in the initial state. Values of 5, 10, 15, 20, and 25 are considered for this parameter in this study.
    
    \item \textbf{Activation function:} An activation function in neural networks produces a value in the output using a node's input values. It maps the weighted sum of the node inputs to a specific domain (depending on the activation function). The final value is transferred to the next layer by this function. In this study, three different activation functions (ReLU, Tanh, and Sigmoid) were investigated (Table~\ref{tab:9}).
    
    \item \textbf{Solver:} Various solvers can be used to adjust the weights and adapt target and output. In this study, three solvers, SGD and Adam, have been used.
    \begin{itemize}
        \item \textbf{SGD (Stochastic Gradient Descent)}
        \item \textbf{AdaDelta:} limits the window of gradients to some fixed size.
        \item \textbf{Adam (Gradient-based Optimizer)}
    \end{itemize}
    The default solver, ``Adam,'' works well in terms of training time and validation score on relatively large datasets (with thousands of training samples or more).
    
    \item \textbf{Alpha coefficient:} Represents the power of data regularization. The loss function is calculated by dividing L2 regularization by sample size. Values of 0.01, 0.05, 0.001, and 0.0001 are considered for this parameter in this study.
    
    \item \textbf{Tolerance:} When a loss or score does not improve over consecutive iterations, convergence is achieved and training stops. Four tolerance values were used in this study: 0.01, 0.001, 0.0001, 0.00001, and 0.000001.
\end{itemize}

\input{Tables/table9}

A total of 702{,}900 models were created by combining different combinations of the hyperparameters. All possible cases are shown in Table~\ref{tab:10}.

\input{Tables/table10}

The optimal model is a neural network with four hidden layers, 10 neurons in the first layer, 15 neurons in the second layer, 20 neurons in the third layer, and 15 neurons in the fourth hidden layer. Tanh function was identified as the activation function, and Adam solver was selected as the proper optimizer, with an alpha coefficient of 0.001 and tolerance of 0.0001. The top five MLP models are presented in Table~\ref{tab:11}.

\input{Tables/table11}

\begin{figure}[h!]
\centering
\begin{subfigure}[b]{0.32\textwidth}
    \includegraphics[width=\textwidth]{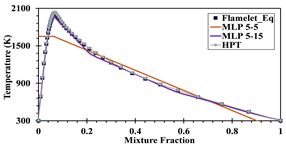}
    \caption{}
    \label{fig:14a}
\end{subfigure}
\hfill
\begin{subfigure}[b]{0.32\textwidth}
    \includegraphics[width=\textwidth]{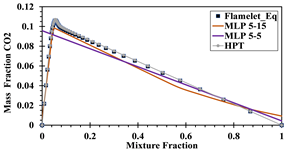}
    \caption{}
    \label{fig:14b}
\end{subfigure}
\hfill
\begin{subfigure}[b]{0.32\textwidth}
    \includegraphics[width=\textwidth]{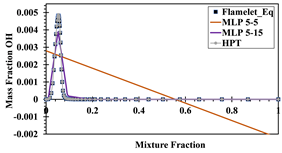}
    \caption{}
    \label{fig:14c}
\end{subfigure}
\caption{\footnotesize Results of the optimal hyperparameters (HPT), the weakest model (MLP$_{5-5}$), and optimal model (MLP$_{5-15}$): (a) Temperature (b) CO$_2$ (c) OH.}
\label{fig:14}
\end{figure}

As a result of the examinations, Figure~\ref{fig:14} presents results of the optimal model using hyperparameter tuning (named HPT) (Table~\ref{tab:11}), the weakest model MLP5-5 (Table~\ref{tab:8}), and the best model MLP5-15 (Table~\ref{tab:8}), based on a scalar dissipation rate of 5~(1/s).

By comparing the data of these three different models with the numerical solution of the flamelet equations, it is concluded that these models can accurately predict different species' temperatures and behavior. According to the model evaluation and the accuracy of the models, there is a small difference between the two models in terms of species (Table~\ref{tab:12}). Hence, the results of MLP5-15 and HPT are similar to those of the flamelet equations. In addition, HPT’s accuracy in predicting other species is acceptable, as shown in Table~\ref{tab:13}.

\input{Tables/table12}

\input{Tables/table13}

\section{Conclusion}

To address the FGM memory and computational time issue, four different machine-learning approaches are implemented in this work. An integrated 1D combustion code is used to test the performance of ML-FGM models using RF, SVR, MLP, and LR. An initial library of cross-flow, with scalar dissipation rates between 0.01 and 29.5, was selected for training ML models and concluded that:

\begin{itemize}
    \item Seven libraries are suitable for creating a machine learning database with an error of 2.30\%.
    
    \item After checking the default architectures of each model, the MLP model with 80.95\% data accuracy was the best model.
    
    \item By considering an MSE value of 0.0005, an accuracy of 90\% is enough for the predictions of MLP to match perfectly with the data extracted from the flamelet equations.
    
    \item Hyperparameter tuning was used to improve the accuracy. Through the simultaneous changing of seven hyperparameters and the combination of these hyperparameters, 702{,}900 different models with varying accuracy were constructed. The model with an accuracy of 99.81\% was selected as the optimal model.
\end{itemize}

\bibliography{bibliography/Bibliography}
\bibliographystyle{IEEEtran}

\end{document}

%% file: Tables/table1.tex
\begin{table}[h!]
\centering
\footnotesize 
\caption{\footnotesize Rules of thumb for choosing proper ML models \cite{yang_application_2022, xie_feature_2023}}
\label{tab:ml_rules}
\renewcommand{\arraystretch}{1.3}
\begin{tabular}{|p{2.5cm}|p{12.5cm}|}
\hline
\textbf{Rule} & \textbf{Description} \\
\hline 
Start with simplicity & In general, simpler models tend to generalize better than complex ones. Before moving on to more complex models, start with simple models such as linear regression. \\
\hline
Consider the size of the dataset & In order to prevent overfitting, choose models with low complexity, if your dataset is small. In contrast, a large dataset can allow you to explore more complex models. \\
\hline
Understand the problem and the data & Gain a deep understanding of the problem and data you are analyzing. You can use this knowledge to select models appropriate for the particular problem domain and data type (for example, classification, regression, structured, or unstructured). \\
\hline
Evaluate model assumptions & A model's assumptions about the underlying data distribution and relationships need to be evaluated. Ensure that your chosen model is aligned with your dataset's assumptions. The linear regression method, for instance, assumes a linear relationship between the features and the target variable. \\
\hline
Accuracy or interpretability & It is important to provide more transparent explanations of predictions, such as linear models or decision trees, if interpretability and understanding are important. The interpretation of complex models like neural networks may be more challenging. \\
\hline
Computational resources & To train and deploy some models, significant computational resources (e.g., memory, processing power) are required. Ensure the models you choose can be implemented on the available computing resources. \\
\hline
\end{tabular}
\end{table}

%% file: Tables/table2.tex
\begin{table}[h!]
\centering
\fontsize{10}{9}
\footnotesize 
\caption{\footnotesize Advantages and disadvantages of applied methods}
\label{tab:2}
\renewcommand{\arraystretch}{1.3}
\begin{tabular}{|p{2.5cm}|p{6.5cm}|p{6.5cm}|}
\hline
\textbf{Model} & \textbf{Advantages} & \textbf{Disadvantages} \\
\hline
Linear regression &
\begin{itemize}
  \item High accuracy
  \item Ease of implementation
\end{itemize} &
\begin{itemize}
  \item Need for hyperparameters such as setting and repetition parameters
  \item Sensitive to feature scaling
\end{itemize} \\
\hline
Random forest &
\begin{itemize}
  \item High versatility
  \item High accuracy and flexibility
  \item Effective way to estimate missing data
\end{itemize} &
\begin{itemize}
  \item Time-consuming to calculate the data for each tree
  \item Requires a lot of computing and storage resources
  \item Possibility of overfitting in a small number of trees
\end{itemize} \\
\hline
Support vector machine &
\begin{itemize}
  \item Effective in large spaces
  \item Effective when dimensions exceed sample size
  \item Versatile due to kernel functions
  \item Ability to specify custom cores
\end{itemize} &
\begin{itemize}
  \item Do not directly provide probability estimates
  \item Computationally expensive
  \item Less interpretable than other algorithms
\end{itemize} \\
\hline
Multilayer perceptron &
\begin{itemize}
  \item Ability to learn nonlinear models
  \item Ability to learn in real time (online learning)
\end{itemize} &
\begin{itemize}
  \item Initial architecture needs to be set
  \item Sensitive to feature scaling
\end{itemize} \\
\hline
\end{tabular}
\end{table}

%% file: Tables/table3.tex
\begin{table}[h!]
\centering
\footnotesize
\caption{\footnotesize Maximum, minimum and average error of each model formed by different number of libraries}
\label{tab:3}
\renewcommand{\arraystretch}{1.3}
\begin{tabular}{|c|cc|cc|cc|}
\hline
\textbf{Number of libraries} & \multicolumn{2}{c|}{\textbf{Maximum error (\%)}} & \multicolumn{2}{c|}{\textbf{Minimum error (\%)}} & \multicolumn{2}{c|}{\textbf{Mean error (\%)}} \\
\cline{2-7}
 & Temperature & CO$_2$ & Temperature & CO$_2$ & Temperature & CO$_2$ \\
\hline
3  & 49.6  & 20.06 & 0.11 & 1.13 & 4.00 & 7.17 \\
7  & 32.19 & 14.24 & 0.04 & 3.59 & 2.30 & 8.82 \\
12 & 28.84 & 19.49 & 0.07 & 2.26 & 2.23 & 9.51 \\
17 & 18.69 & 24.94 & 1.43 & 9.15 & 2.06 & 15.08 \\
22 & 20.19 & 14.04 & 0.02 & 1.62 & 1.69 & 6.55 \\
\hline
\end{tabular}
\end{table}

%% file: Tables/table4.tex
\begin{table}[h!]
\centering
\footnotesize
\caption {\footnotesize Seven libraries and scalar dissipation rates}
\label{tab:4}
\renewcommand{\arraystretch}{1.3}
\begin{tabular}{|c|c|c|c|c|c|c|c|}
\hline
\textbf{Scalar Dissipation Rate} & $\chi_1$ & $\chi_2$ & $\chi_3$ & $\chi_4$ & $\chi_5$ & $\chi_6$ & $\chi_7$ \\
\hline
\textbf{Value} & 0.01 & 5.5 & 10 & 14.5 & 20.5 & 25 & 29.5 \\
\hline
\end{tabular}
\end{table}

%% file: Tables/table5.tex
\begin{table}[h!]
\centering
\footnotesize
\caption{\footnotesize Five different models with different prediction accuracies in $\chi=5$ (1/s)}
\label{tab:5}
\renewcommand{\arraystretch}{1.3}
\begin{tabular}{|c|c|c|ccc|}
\hline
\textbf{Model} & \textbf{Architecture} & \textbf{Accuracy of the model} & \multicolumn{3}{c|}{\textbf{Accuracy compared to Flamelet equations}} \\
\cline{4-6}
& & & Temperature & CO$_2$ & OH \\
\hline
1 & MLP$_{3-15}^{a}$ & 96.77 & 95.81 & 92.48 & 93.65 \\
2 & MLP$_{5-10}$     & 91.32 & 91.50 & 84.57 & 89.98 \\
3 & MLP$_{2-10}$     & 86.93 & 84.16 & 76.43 & 88.12 \\
4 & MLP$_{1-20}$     & 62.28 & 60.45 & 55.60 & 52.05 \\
5 & MLP$_{5-5}$      & 44.90 & 50.21 & 40.18 & 35.62 \\
\hline
\end{tabular}

\vspace{2mm}
\raggedright
\footnotesize $^{a}$ MLP (Multi-Layer Perceptron), 3 is the number of hidden layers, and 15 is the number of neurons in each layer.
\end{table}

%% file: Tables/table6.tex
\begin{table}[h!]
\centering
\tiny
\caption{\footnotesize Comparison of temperature (K) and mass fraction of CO$_2$ in 9 different mixture fractions.}
\label{tab:6}
\setlength{\tabcolsep}{1.5pt}
\renewcommand{\arraystretch}{1.8}

\begin{tabular}{|c|cccccc|cccccc|}
\hline
{\textbf{Mixture Fraction}} & \multicolumn{6}{c|}{\textbf{Temperature (K)}} & \multicolumn{6}{c|}{\textbf{CO$_2$ Mass Fraction}} \\
\cline{2-13}
 & \textbf{Flamelet Eq} & \textbf{Model 1} & \textbf{Model 2} & \textbf{Model 3} & \textbf{Model 4} & \textbf{Model 5} 
 & \textbf{Flamelet Eq} & \textbf{Model 1} & \textbf{Model 2} & \textbf{Model 3} & \textbf{Model 4} & \textbf{Model 5} \\
\hline
0.106 & 1820 & 1787 & 1787 & 1805 & 1729 & 1700 & 0.0925 & 0.0946 & 0.0955 & 0.0980 & 0.0951 & 0.0917 \\
0.21  & 1450 & 1452 & 1449 & 1591 & 1542 & 1519 & 0.0840 & 0.0826 & 0.0826 & 0.0862 & 0.0840 & 0.0820 \\
0.30  & 1230 & 1220 & 1196 & 1406 & 1381 & 1363 & 0.0745 & 0.0727 & 0.0735 & 0.0759 & 0.0743 & 0.0735 \\
0.387 & 1060 & 1035 & 1096 & 1227 & 1225 & 1212 & 0.0653 & 0.0646 & 0.0648 & 0.0661 & 0.0650 & 0.0653 \\
0.503 &  879 &  881 &  828 &  989 & 1016 & 1011 & 0.0529 & 0.0523 & 0.0528 & 0.0529 & 0.0526 & 0.0545 \\
0.659 &  678 &  673 &  651 &  668 &  737 &  740 & 0.0363 & 0.0360 & 0.0364 & 0.0352 & 0.0359 & 0.0396 \\
0.757 &  567 &  543 &  563 &  526 &  561 &  570 & 0.0259 & 0.0258 & 0.0259 & 0.0240 & 0.0251 & 0.0306 \\
0.869 &  443 &  440 &  462 &  426 &  400 &  375 & 0.0139 & 0.0142 & 0.0144 & 0.0113 & 0.0131 & 0.0201 \\
1.0   &  300 &  319 &  340 &  300 &  125 &  148 & 0.0000 & 0.0009 & 0.0016 & 0.0004 & 0.0004 & 0.0078 \\
\hline
\end{tabular}
\end{table}

%% file: Tables/table7.tex
\begin{table}[h!]
\centering
\footnotesize
\caption{\footnotesize Dividing and coloring accuracy and MSE in prediction.}
\label{tab:7}
\renewcommand{\arraystretch}{1.3}
\begin{tabular}{cccccc}
\hline
\textbf{Accuracy} 
& $X < 50$ 
& $50 < X < 60$ 
& $60 < X < 80$ 
& $80 < X < 90$ 
& $90 < X$ \\

\textbf{MSE} 
& $Y < 10^{-2}$ 
& $0.01 < Y < 0.005$ 
& $0.005 < Y < 0.001$ 
& $0.001 < Y < 0.0005$ 
& $0 < Y < 0.0005$ \\
\hline
\end{tabular}
\end{table}

%% file: Tables/table8.tex
\begin{table}[h!]
\centering
\caption{\footnotesize Comparison of accuracy, error, training and prediction time of uniform architectures of MLP.}
\label{tab:8}
\footnotesize
\renewcommand{\arraystretch}{1.2}

\begin{tabular}{
ccccccc
}
\hline
\textbf{Number of} & \textbf{Neurons} & \textbf{Name} & \textbf{Accuracy} & \textbf{MSE} & \textbf{Training time} & \textbf{Prediction time} \\
\textbf{hidden layers} & & & (\%) & &  (s) &  (s) \\
\hline
5 & 25 & MLP$_{5-25}^a$ & 98.64 & 0.0005 & 14.25 & 1.96 \\
5 & 20 & MLP$_{5-20}$ & 96.94 & 0.0011 & 15.13 & 2.43 \\
5 & 15 & MLP$_{5-15}$ & 99.34 & 0.0003 & 10.61 & 1.89 \\
5 & 10 & MLP$_{5-10}$ & 91.32 & 0.0038 & 12.52 & 1.57 \\
5 & 5  & MLP$_{5-5}$  & 44.90 & 0.0301 & 2.65  & 0.36 \\
\hline
{4} & 25 & MLP$_{4-25}$ & 98.35 & 0.0009 & 21.27 & 1.64 \\
4 & 20 & MLP$_{4-20}$ & 98.67 & 0.0005 & 12.65 & 2.01 \\
 4& 15 & MLP$_{4-15}$ & 97.45 & 0.0010 & 15.39 & 1.86 \\
 4& 10 & MLP$_{4-10}$ & 94.49 & 0.0021 & 23.17 & 1.73 \\
 4 & 5  & MLP$_{4-5}$  & 73.54 & 0.0102 & 3.78  & 0.27 \\
\hline
{3} & 25 & MLP$_{3-25}$ & 98.20 & 0.0010 & 17.65 & 1.24 \\
3 & 20 & MLP$_{3-20}$ & 97.94 & 0.0009 & 11.33 & 1.78 \\
3 & 15 & MLP$_{3-15}$ & 96.77 & 0.0013 & 12.15 & 1.60 \\
3 & 10 & MLP$_{3-10}$ & 90.36 & 0.0039 & 31.55 & 1.17 \\
3 & 5  & MLP$_{3-5}$  & 74.76 & 0.0097 & 3.10  & 0.20 \\
\hline
{2} & 25 & MLP$_{2-25}$ & 94.51 & 0.0012 & 15.24 & 1.34 \\
2 & 20 & MLP$_{2-20}$ & 96.31 & 0.0014 & 16.51 & 1.29 \\
2 & 15 & MLP$_{2-15}$ & 98.16 & 0.0007 & 14.41 & 1.15 \\
2 & 10 & MLP$_{2-10}$ & 86.93 & 0.0061 & 17.45 & 0.77 \\
2 & 5  & MLP$_{2-5}$  & 88.95 & 0.0051 & 13.59 & 0.51 \\
\hline
{1} & 25 & MLP$_{1-25}$ & 91.80 & 0.0019 & 24.20 & 0.57 \\
1 & 20 & MLP$_{1-20}$ & 93.58 & 0.0030 & 27.13 & 0.66 \\
1 & 15 & MLP$_{1-15}$ & 94.29 & 0.0025 & 28.37 & 0.58 \\
1 & 10 & MLP$_{1-10}$ & 69.05 & 0.0256 & 16.30 & 0.40 \\
1 & 5  & MLP$_{1-5}$  & 62.28 & 0.0245 & 19.47 & 0.31 \\
\hline
{\shortstack{Other\\Models}} 
 & 100 (default) & MLP$_\text{def}$ & 80.95 & 0.0162 & 3.16 & 1.40 \\
 & - & RFR & 96.50 & 0.0016 & 14.06 & 1.82 \\
 & - & LR  & 0.519 & 0.0222 & 19.41 & 0.95 \\
\hline
\end{tabular}

\vspace{0.3em}
{\footnotesize $^a$ MLP: Multi-Layer Perceptron, 5 is number of hidden layers and 25 is number of neurons in each layer}
\end{table}

%% file: Tables/table9.tex
\begin{table}[h!]
\centering
\footnotesize
\caption{\footnotesize Activation functions transfer domain and equations.}
\label{tab:9}
\renewcommand{\arraystretch}{1.3}
\begin{tabular}{|c|c|c|}
\hline
\textbf{Activation function} & \textbf{Transfer domain} & \textbf{Equation} \\
\hline
ReLU & $(0, \infty)$ & $\varphi(x) = \max(0, z)$ \\
Tanh & $(-1, 1)$ & $\varphi(x) = \frac{e^z - e^{-z}}{e^z + e^{-z}}$ \\
Sigmoid & $(0, 1)$ & $\varphi(x) = \frac{1}{1 + e^{-z}}$ \\
\hline
\end{tabular}
\end{table}

%% file: Tables/table10.tex
\begin{table}[h!]
\centering
\footnotesize
\caption{\footnotesize All the hyperparameters.}
\label{tab:10}

\renewcommand{\arraystretch}{1.1}

\begin{tabular}{|c|c|c|c|c|c|}
\hline
\textbf{Hidden Layers} & \textbf{ Neurons} & \textbf{Activation Function} & \textbf{Solver} & \textbf{Alpha coefficient} & \textbf{Tolerance} \\
\hline
1 & 5  & Sigmoid & SGD      & 0.01   & 0.01 \\
2 & 10 &         &          & 0.05   & 0.001 \\
3 & 15 & ReLU    & Adam     & 0.001  & 0.0001 \\
4 & 20 &         &          & 0.001  & 0.00001 \\
5 & 25 & Tanh    & AdaDelta & 0.0001 & 0.000001 \\
\hline
\end{tabular}
\end{table}

%% file: Tables/table11.tex
\begin{table}[h!]
\centering
\footnotesize
\caption{\footnotesize Top five MLP models using hyperparameter tuning}
\label{tab:11}
\renewcommand{\arraystretch}{1.3}
\begin{tabular}{|c|c|c|c|c|c|c|c|}
\hline
\textbf{Fittest} & \textbf{No. Hidden layers} & \textbf{Configuration} & \textbf{Function} & \textbf{Solver} & \textbf{Alpha} & \textbf{Tol} & \textbf{Accuracy} \\
\hline
1\textsuperscript{st} & 4 & 10-15-20-15          &  &  &  &  & 99.81 \\
2\textsuperscript{nd} & 4 & 10-15-15-15          &                        &                        &                         &                          & 99.76 \\
3\textsuperscript{rd} & 5 & 15-15-15-15-15        &      {Tanh}                  &      {Adam}                  &     0.001                    &   0.00001                       & 99.34 \\
4\textsuperscript{th} & 5 & 10-10-15-10-15        &                        &                        &                         &                          & 99.32 \\
5\textsuperscript{th} & 4 & 15-15-20-15           &                        &                        &                         &                          & 99.21 \\
\hline
\end{tabular}
\end{table}

%% file: Tables/table12.tex
\begin{table}[h!]
\centering
\footnotesize
\caption{\footnotesize Temperature and species prediction accuracy of HPT, MLP\textsubscript{5-15} and MLP\textsubscript{5-5}}
\label{tab:12}
\renewcommand{\arraystretch}{1.3}
\begin{tabular}{|c|c|c|c|}
\hline
\textbf{Model} & \textbf{Temperature accuracy} & \textbf{CO$_2$ mass fraction accuracy} & \textbf{OH mass fraction accuracy} \\
\hline
HPT & 99.62 & 99.25 & 98.67 \\
MLP\textsubscript{5-15} & 99.48 & 98.92 & 97.58 \\
MLP\textsubscript{5-5}  & 50.21 & 40.18 & 35.62 \\
\hline
\end{tabular}
\end{table}

%% file: Tables/table13.tex
\begin{table}[h!]
\centering
\footnotesize
\caption{\footnotesize HPT’s accuracy on predicting other species (\%).}
\label{tab:13}
\renewcommand{\arraystretch}{1.3}
\begin{tabular}{|c|c|c|c|c|c|}
\hline
\textbf{Species} & \textbf{Accuracy} & \textbf{Species} & \textbf{Accuracy} & \textbf{Species} & \textbf{Accuracy} \\
\hline
H$_2$     & 97.70 & CH$_3$     & 96.37 & C$_2$H$_5$ & 94.95 \\
H         & 93.73 & CH$_4$     & 98.70 & C$_2$H$_6$ & 96.68 \\
O         & 96.58 & CO         & 99.36 & HCCO       & 93.91 \\
O$_2$     & 97.00 & HCO        & 96.15 & CH$_2$CO   & 94.14 \\
OH        & 92.66 & CH$_2$O    & 96.43 & HCCOH      & 92.28 \\
H$_2$O    & 97.02 & CH$_2$OH   & 92.72 & N          & 92.86 \\
HO$_2$    & 94.36 & CH$_3$O    & 94.28 & NH         & 98.63 \\
H$_2$O$_2$& 96.83 & CH$_3$OH   & 97.91 & NH$_2$     & 97.63 \\
C         & 97.90 & C$_2$H     & 92.90 & NH$_3$     & 97.61 \\
CH        & 95.34 & C$_2$H$_2$ & 92.05 & NNH        & 99.13 \\
CH$_2$    & 97.70 & C$_2$H$_3$ & 96.37 & NO         & 94.95 \\
CH$_2$(s) & 93.73 & C$_2$H$_4$ & 98.70 & NO$_2$     & 96.68 \\
N$_2$O    & 94.40 & HNO        & 93.59 & CN         & 94.08 \\
HCN       & 96.13 & H$_2$CN    & 98.32 & HCNN       & 92.63 \\
HCNO      & 93.15 & HOCN       & 96.39 & HNCO       & 96.72 \\
NCO       & 95.52 & CH$_2$CHO  & 94.40 & N$_2$      & 93.59 \\
\hline
\end{tabular}
\end{table}